\pdfoutput=1

\documentclass[11pt]{article}
\usepackage[final]{acl}
\usepackage{times}
\usepackage{latexsym}
\usepackage{makecell}
\usepackage[T1]{fontenc}

\usepackage[utf8]{inputenc}

\usepackage{microtype}

\usepackage{placeins}

\usepackage{tcolorbox}
\usepackage{inconsolata}
\usepackage{booktabs}
\usepackage{tabularx}
\usepackage{array}
\usepackage{ragged2e}
\usepackage{caption}
\usepackage{graphicx}

\title{Answer-Centric or Reasoning-Driven? Uncovering the Latent Memory Anchor in LLMs}

\author{
 \textbf{Yang Wu\textsuperscript{1,2}},
 \textbf{Yifan Zhang\textsuperscript{1,2,6}$^{*}$},
 \textbf{Yiwei Wang\textsuperscript{3}},
 \textbf{Yujun Cai\textsuperscript{4}},
\\
 \textbf{Yurong Wu\textsuperscript{2}},
 \textbf{Yuran Wang\textsuperscript{5}},
 \textbf{Ning Xu\textsuperscript{1}},
 \textbf{Jian Cheng \textsuperscript{1,2,6}}
\\
\\
 \textsuperscript{1}C$^2$DL, Institute of Automation, Chinese Academy of Sciences \\
  \textsuperscript{2}School of Artificial Intelligence, University of Chinese Academy of Sciences, Beijing\\
 \textsuperscript{3}UC Merced,
  \textsuperscript{4}The University of Queensland, 
   \textsuperscript{5}Dalian Minzu University\\
 \textsuperscript{6}University of Chinese Academy of Sciences, Nanjing
\\
{
   \textbf{Correspondence:} {yfzhang@nlpr.ia.ac.cn}
}
}

\newcommand{\AEfull}{\textbf{Answer-Explicit (AE)}}
\newcommand{\AERfull}{\textbf{Answer-Embedded-Reasoning (AER)}}
\newcommand{\AMRfull}{\textbf{Answer-Masked-Reasoning (AMR)}}
\newcommand{\ARRfull}{\textbf{Answer-Removed-Reasoning (ARR)}}
\newcommand{\AFfull}{\textbf{Answer-Free (AF)}}

\newcommand{\deepseek}{Deepseek-R1~\citep{guo2025deepseek}}
\newcommand{\oThree}{OpenAI-o3~\citep{IntroducingOpenAIO3}}
\newcommand{\oThreeMini}{OpenAI-o3-mini-high~\citep{OpenAIO3mini}}
\newcommand{\oFourMini}{OpenAI-o4-mini-high~\citep{IntroducingOpenAIO3}} 
\newcommand{\qwq}{QWQ-32B~\citep{teamQwQ32BEmbracingPower2025}}
\newcommand{\grokBeta}{Grok3-Mini-Beta-high~\citep{Grok3Beta}}
\newcommand{\grokFast}{Grok3-Mini-Fast-Beta-high~\citep{Grok3Beta}}
\newcommand{\geminiFlash}{Gemini-2.5 Flash Preview-0417~\citep{DevelopersCanNow2025}}
\newcommand{\geminiFlashnew}{Gemini-2.5 Flash Preview-\\0417~\citep{DevelopersCanNow2025}}
\newcommand{\geminiPro}{Gemini-2.5 Pro Preview~\citep{Gemini25Our2025}}
\newcommand{\claude}{Claude 3.7 Sonnet~\citep{anthropic2025claude37}}

\begin{document}
\maketitle
\begin{abstract}
While Large Language Models (LLMs) demonstrate impressive reasoning capabilities, growing evidence suggests much of their success stems from memorized answer-reasoning patterns rather than genuine inference. In this work, we investigate a central question: are LLMs primarily anchored to final answers or to the textual pattern of reasoning chains?  We propose a five-level answer-visibility prompt framework that systematically manipulates answer cues and probes model behavior through indirect, behavioral analysis.  Experiments across state-of-the-art LLMs reveal a strong and consistent reliance on explicit answers. The performance drops by 26.90\% when answer cues are masked, even with complete reasoning chains. These findings suggest that much of the reasoning exhibited by LLMs may reflect post-hoc rationalization rather than true inference, calling into question their inferential depth. Our study uncovers the answer-anchoring phenomenon with rigorous empirical validation and underscores the need for a more nuanced understanding of what constitutes reasoning in LLMs.
\end{abstract}

\section{Introduction}\label{sec:intro}\
In recent years, Large Language Models (LLMs) enhanced by Chain‑of‑Thought (CoT) \citep{wei2022chain} have achieved remarkable success across tasks such as code generation and natural language understanding \citep{zhao2023survey, jiang2024survey, chang2024survey}. Within this progress, mathematical reasoning \citep{ahn2024large} has emerged as a definitive proving ground, with models generating multi-step solutions spanning from elementary arithmetic to graduate-level proofs. However, recent empirical evidence indicates that many of these reasoning chains stem not from genuine inference, but from the recitation of answer-reasoning patterns memorized during training~\citep{xie2024memorization, yan2025recitation}.

\begin{figure*}[ht]
\begin{center}
\includegraphics[width=1\textwidth, trim=0cm 0cm 0cm 0cm, clip]{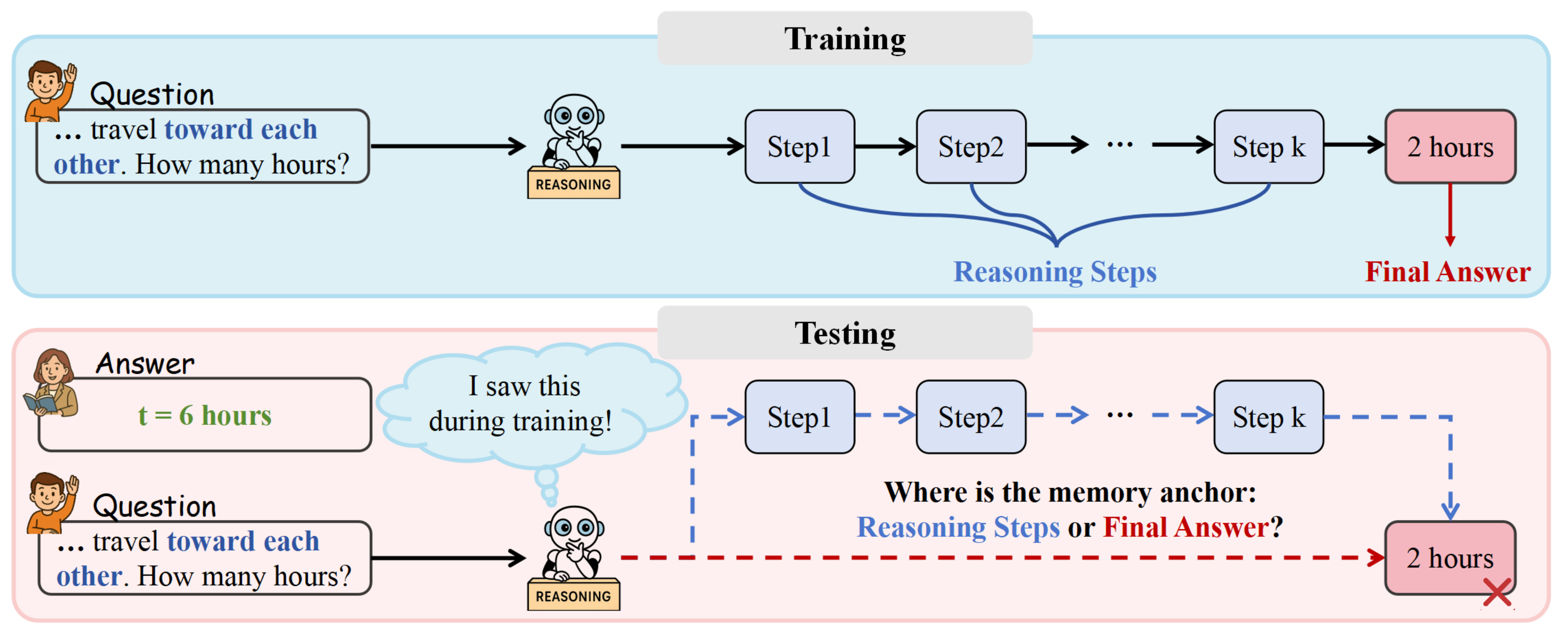}
\end{center}

\caption{When confronted with an unseen yet similar problem, an LLM often recalls a memorized answer–reasoning pattern, overlooking task-specific nuances and producing an incorrect output. The underlying cause of this reliance remains unclear: is the recall anchored to the final answer or to the reasoning chain?}
\label{fig:intro}
\end{figure*}

This pattern of memorization can be exacerbated by the widespread data contamination in large-scale pretraining corpora~\citep{li2023open, xu2024benchmark, chen2025recent}. This memorization inflates headline accuracy while obscuring the absence of genuine inference, because each pattern binds the final answer to a pre-written reasoning chain. Although the shortcut boosts performance on familiar problems, it also enlarges the training–testing gap~\citep{xie2023adaptive, kang2024learning}. As illustrated in Figure~\ref{fig:intro},  even minor input perturbations can induce brittle, systematic failures—reflecting behavior anchored to surface-level patterns rather than flexible, abstract reasoning. Developing such deeper reasoning remains difficult, since current training methods primarily reward fitting to data rather than fostering generalizable reasoning skills.

This reliance on learned patterns restricts models from truly grasping the logic of novel problems, highlighting a critical and underexplored question:

\textit{Are LLMs primarily anchored to final answers or to learned solution templates—namely, the textual patterns of their reasoning chains?}

In this work, we aim to address this fundamental question. A central challenge is the opaque nature of modern LLMs. These models largely operate as "black boxes"~\citep{zhao2024explainability, yang2025challenge}, limiting direct visibility into their internal memory traces. Consequently, it remains unclear what these models have memorized or how this information is utilized during inference. To overcome this limitation, we employ an indirect, behavioral methodology. Specifically, our approach involves systematically manipulating the visibility and form of final answers and reasoning steps within model inputs. By analyzing how these manipulations impact model outputs, we infer whether model behavior is primarily driven by memorized answers or adherence to learned solution templates, thereby illuminating the latent “memory anchor” that guides LLM responses.

To operationalize this investigation, our methodology centers on designing a systematic series of prompts for CoT reasoning. These prompts precisely vary the explicitness of final answers within reasoning chains, creating a spectrum of conditions ranging from fully visible answers, through partial answer masking, to the complete removal of answer cues. Such controlled manipulations, while holding other prompt variables constant, enable us to rigorously quantify performance changes and thereby assess the model's dependency.

Our experiments consistently reveal a striking trend: models with explicit access to correct answers demonstrate substantially improved performance, often achieving near-perfect accuracy. In contrast, once the final answer is methodically obscured or answer-containing sentences are removed—even when the complete reasoning structure is preserved—performance declines markedly. This pronounced performance gradient strongly indicates that LLMs' memories are predominantly anchored to answers rather than the textual patterns of their reasoning chains.

The implications of our findings are considerable and contribute meaningfully to ongoing scholarly debates regarding the authenticity of reasoning in LLMs. Our results suggest that the widely-used CoT reasoning patterns generated by LLMs may often function as post-hoc rationalizations~\citep{arcuschin2025chain} rather than reflecting genuine inferential steps. This reliance on answers further compromises the models' robustness and significantly impairs their generalization ability, especially when encountering out-of-distribution or subtly perturbed  problems. 

This study makes three primary contributions:
\begin{itemize}
    \item Present comprehensive evidence across state‑of‑the‑art LLMs showing that answer anchoring, rather than reasoning-template recall, is the dominant memory mechanism.

    \item  Introduce a five-level prompt framework that isolates answer cues from reasoning chains to diagnose memory anchoring behavior.

    \item  Demonstrate the tenacity of answer anchoring via conflict and warning-based prompts that test LLMs’ resistance to answer cue overrides.
\end{itemize}

\section{Related Work}\label{sec:relate}\

\subsection{Memorization and Reasoning of LLMs}

Reasoning is widely recognized as a key capability of LLMs. However, recent work challenges this view, suggesting that much of the performance attributed to reasoning may instead stem from memorizing training data patterns~\citep{xie2024memorization,  jiang2024memorization, qiu2024can, chen2025memorize}. These findings call into question how often LLMs reason rather than retrieve.

Studies in the mathematical domain particularly underscore this concern. Across diverse tasks—including noisy rule induction, subtly rephrased problems, and logic puzzles—LLMs consistently tend to rely on memorized solution templates~\citep{li2025patterns, huang2025math, xie2024memorization}. \citep{yan2025recitation}  further substantiated these observations. Their work demonstrates that even high-performing LLMs often recite templates for elementary problems and exhibit significant performance degradation under minor perturbations. Our work offers a finer-grained perspective on this memorization.  We investigate whether LLMs' recall is primarily anchored to final answers or to the textual structure of reasoning chains, aiming for deeper mechanistic insights.

\subsection{Behavioral Probing via Input Manipulation}
The inherent black-box~\citep{cheng2024black, zhao2024explainability, yang2025challenge} nature of LLMs makes direct inspection of their internal mechanism impractical. Consequently, behavioral probing via input manipulation has emerged as a key strategy to understand these opaque systems. This approach involves systematically altering model inputs and observing the resultant output changes to infer underlying processes. Prior work has explored a variety of such manipulations. Some studies focus on numerical and linguistic perturbations~\citep{li2024gsm, zhou2024can, shrestha2025mathematical, chatziveroglou2025exploring}. Others  adjust element order~\citep{pezeshkpour2024large, chen2024premise, guan2025order} or employ masking and deletion of key content~\citep{wu2024large, chen2024masked, fan2025missing}. Furthermore, some evaluations integrate multiple manipulation strategies~\citep{liang2022holistic, zhu2023promptrobust}. Distinctly, our work establishes an admittedly artificial yet diagnostically critical scenario: providing the correct final answer with the input problem. This approach leverages LLMs' post-hoc rationalization to generate high-quality reasoning chains, thereby creating unique conditions to compare the impacts of explicit answers versus the reasoning chains on model performance.

\begin{figure*}[ht]
\begin{center}
\includegraphics[width=1.0\textwidth, trim=0cm 0cm 0cm 0cm, clip]{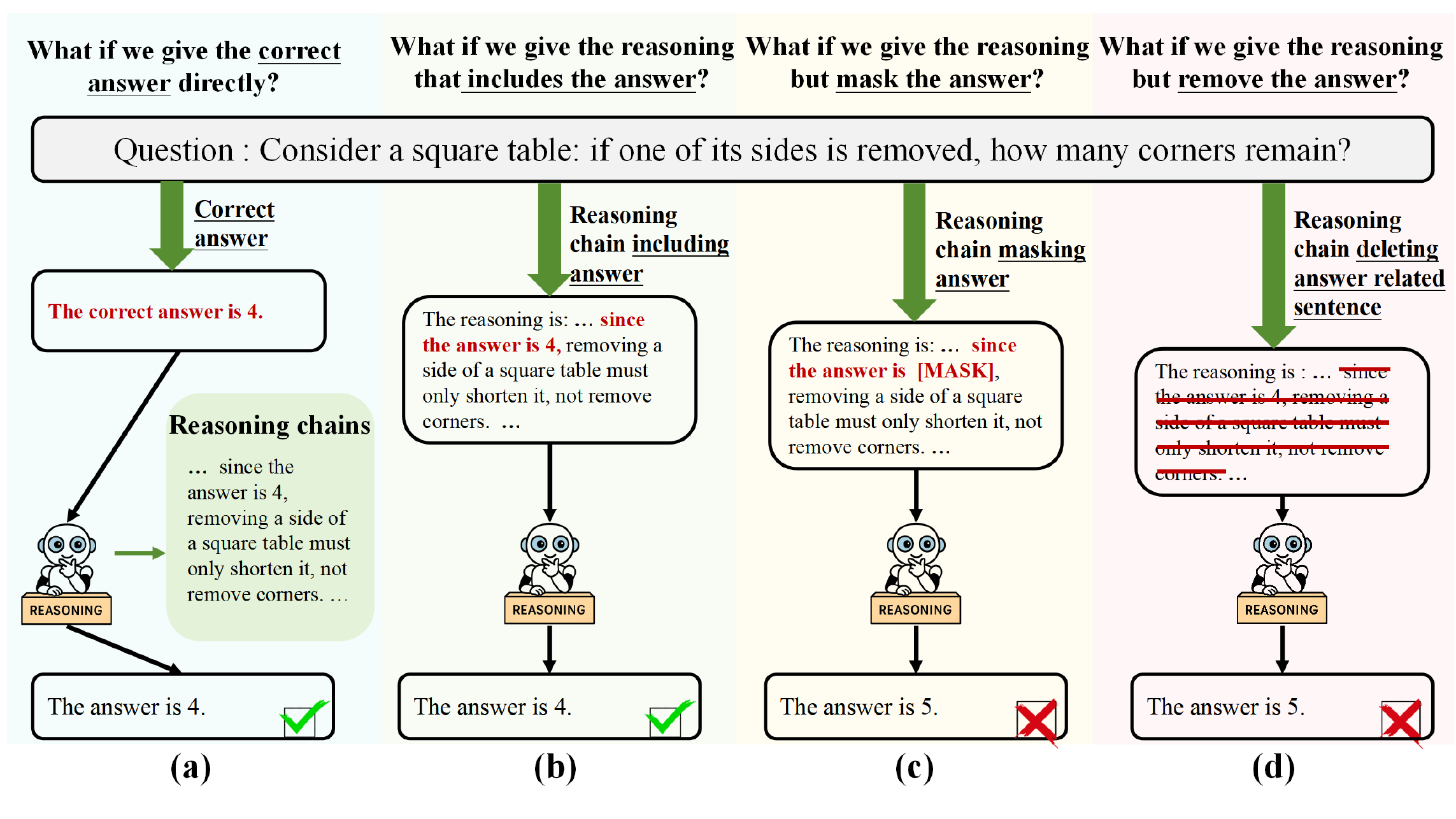}
\end{center}

\caption{Schematic of LLMs responses to a reasoning task under systematically manipulated input prompts that vary answer visibility. The model predicts correctly when the answer is explicitly provided, either directly (a) or embedded within the reasoning chain (b). However, performance sharply declines when these explicit answer cues are obscured by masking (c) or by removing answer-related sentences from the reasoning chain (d), highlighting a strong dependency on readily available answers.
}
\label{fig:framework}
\end{figure*}

\section{Investigating Memory Binding in Reasoning Models}\label{sec:method}

This section details our methodology for investigating whether LLMs primarily anchor memory to final answers or to learned solution templates. We employ an indirect approach, systematically manipulating input elements related to answer and reasoning visibility (as illustrated in Figure~\ref{fig:framework}) to infer underlying model dependencies.

The section is organized as follows: we first present a motivating example (Section~\ref{subsec:motivation}), then detail our specific hypothesis and experimental design (Section~\ref{subsec:hypothesis}), and finally outline the experimental setup including datasets, models, and evaluation metrics (Section~\ref{subsec:setup}).

\subsection{Illustrative Example: Dependency on Answer Visibility}\label{subsec:motivation}

Consider a simple reasoning problem, illustrated in Figure~\ref{fig:framework}, that asks whether removing one side of a square alters its corner count. Despite its simplicity, the models’ predictions vary dramatically with how answer cues are presented in the prompt. The models answer correctly (e.g., "four corners") when the solution is explicit ((a) Answer-Explicit or (b) Answer-Embedded-Reasoning). However, performance sharply deteriorates when these explicit cues are obscured.  For example, masking the answer token (c) or removing answer-relevant sentences (d) typically leads to failure. In such cases, the models might incorrectly predict "five corners".

This stark contrast underscores the models' pronounced dependence on explicit answers, even on such an elementary task. Their sensitivity to readily available answers appears to overshadow the reasoning process itself. Such behavior directly motivates our central investigation: are these models primarily recalling memorized answers, or are they merely following the textual patterns of provided reasoning chains (i.e., solution templates)?

\subsection{Hypothesis and Experimental designs}\label{subsec:hypothesis}

Motivated by the behavioral patterns observed in Section \ref{subsec:motivation}, we hypothesize that LLMs rely more strongly on explicit final-answer cues than on learned solution templates.

Directly verifying this hypothesis via internal state inspection is generally infeasible due to the black-box nature of LLMs. We therefore adopt an indirect, prompt-based intervention framework to probe this dependency. This framework involves systematically manipulating the explicitness of final answer cues in prompts and observing the corresponding impact on model performance. Through comparative analysis across these controlled variations, we can infer the primary anchor of the model's memory binding.

To operationalize this, we define five prompt conditions that create a graduated spectrum of answer cue visibility:

\AEfull: The prompt provides the problem and its correct final answer explicitly (Figure~\ref{fig:framework}a), offering maximal answer-cue visibility.

\AERfull: The prompt provides the correct answer, but it is embedded within a full reasoning chain derived from the AE prompt (Figure~\ref{fig:framework}b).

\AMRfull: The prompt provides the full reasoning chain, but every occurrence of the final answer is replaced by a placeholder (e.g., \texttt{[MASK]}), while sentences hinting at the answer may remain (Figure~\ref{fig:framework}c).

\ARRfull: The prompt provides a pruned reasoning chain in which any sentence or clause that directly states—or unmistakably reveals—the final answer has been completely removed (Figure~\ref{fig:framework}d).

\AFfull: The prompt provides only the problem statement, without any supplementary answer or reasoning cues.

This systematic design, which ranges from explicit answer provision (AE, AER) to the almost complete removal of answer cues within the reasoning chain (AMR, ARR, AF), allows a precise assessment of the model’s sensitivity to answer visibility. Performance comparisons across these variants reveal whether successful predictions arise mainly from memorized final answers or from engagement with learned solution templates. A steep drop in accuracy as answer cues fade would indicate a stronger dependence on those answers than on the reasoning template itself.

\subsection{Experimental Setup}\label{subsec:setup}

\textbf{Dataset}
Our experiments are conducted on the text-only subset of RoR-Bench (Recitation-over-Reasoning Benchmark)~\citep{yan2025recitation}, which is designed to evaluate the robustness of LLM reasoning under subtle input perturbations. RoR-Bench is constructed by applying controlled edits to 158 original Chinese questions spanning arithmetic, logic, optimization, and commonsense reasoning. We evaluate models on these edited Chinese prompts to preserve the benchmark’s intent. For clarity, illustrative examples from the benchmark in this paper are translated into English. 

The RoR-Bench problems are well-suited to our prompt-intervention framework for several reasons. First, they are concise and unambiguous. Second, they feature minimal surface complexity with clearly defined final answers. Finally, they have a low likelihood of training-data contamination.

While RoR-Bench also includes a set of visual problems, we leave the investigation of answer-reasoning dependence in multi-modal models to future work.

\textbf{Prompt Design} For each problem, we generate five distinct prompt variants, corresponding to the conditions in Section~\ref{subsec:hypothesis}. Our central manipulation targets the presentation of answer-related content, while all other prompt elements, such as the problem description and overall format, are kept fixed. This controlled approach ensures that observed differences in model behavior are attributable solely to how answer information is provided.

In the AMR condition, each answer phrase in the chain is automatically replaced with a \texttt{[MASK]} placeholder using GPT-4o-1120 \citep{hurst2024gpt4o}. This process ensures contextually appropriate and consistent substitutions (see Appendix~\ref{subsec:mask_prompt} for masking details).

\textbf{Models} We evaluate a set of models with long thinking process: \deepseek, \oThree, \oThreeMini, \oFourMini, \qwq, \grokBeta, \grokFast, \geminiFlash, \geminiPro, and \claude.

A key distinction in our experimental protocol relates to the models' ability to expose intermediate reasoning steps. Models such as Deepseek-R1, Grok, and QWQ-32B offer this visibility and are thus evaluated under all five prompt conditions. In contrast, models from the OpenAI and Gemini families provide no such readily accessible traces, so we evaluate them only in the AE and AF settings, which do not necessitate inspection of those traces.

All models are prompted in a zero-shot setting. To ensure deterministic outputs, the temperature parameter is uniformly set to 0 for all experiments.

\textbf{Evaluation} We evaluate model-generated responses for answer accuracy by assigning a binary score. A response receives 1 if it exactly matches the ground-truth answer, and 0 otherwise. All outputs are scored automatically. Following the evaluation protocol established by \citet{yan2025recitation}, we employ GPT-4o-1120 as an automated verifier. Details of this verifier, including the specific prompt, are provided in Appendix~\ref{subsec:judge}. This protocol is designed to quantify how model performance varies under the five prompt conditions.

\begin{table*}[htbp]
\small
\centering
\caption{Performance comparison of various reasoning models across different prompt conditions.}
\label{tab:performance_comparison}
\setlength{\tabcolsep}{3.3pt}
\begin{tabular}{lccccc}
\toprule
\textbf{Models} & \textbf{\makecell[c]{Answer\\Explicit}} & \textbf{\makecell[c]{Answer-Embedded\\Reasoning}} & \textbf{\makecell[c]{Answer-Masked\\Reasoning}} & \textbf{\makecell[c]{Answer-Removed\\Reasoning}} & \textbf{\makecell[c]{Answer\\Free}} \\
\midrule
\deepseek & 93.04\% & 84.18\% & 55.06\% & 46.84\% & 23.42\% \\
\oThree & 86.08\% & -- & -- & -- & 25.95\% \\
\oThreeMini & 92.41\% & -- & -- & -- & 25.95\% \\
\oFourMini & 79.11\% & -- & -- & -- & 28.48\% \\
\qwq & 85.44\% & 81.65\% & 58.23\% & 40.51\% & 25.32\% \\
\grokBeta & 86.71\% & 70.25\% & 59.49\% & 50.63\% & 33.54\% \\
\grokFast & 88.61\% & 71.52\% & 60.76\% & 52.53\% & 32.28\% \\
\makecell[l]{\geminiFlashnew} & 72.15\% & -- & -- & -- & 27.22\% \\
\geminiPro & 70.25\% & -- & -- & -- & 31.01\% \\
\claude & 93.67\% & 77.22\% & 55.70\% & 41.77\% & 28.48\% \\
\midrule
\textbf{Avg. Performance} & 
\makecell[c]{\textbf{84.75\%} \\ ($\pm$8.36\%)} & 
\makecell[c]{\textbf{76.96\%} \\ ($\pm$6.10\%)} & 
\makecell[c]{\textbf{57.85\%} \\ ($\pm$2.43\%)} & 
\makecell[c]{\textbf{46.46\%} \\ ($\pm$5.29\%)} & 
\makecell[c]{\textbf{28.17\%} \\ ($\pm$3.26\%)} \\
\textbf{Avg. Decrease} & N/A & \textbf{-7.79\%} & \textbf{-26.90\%} & \textbf{-38.29\%} & \textbf{-56.58\%} \\
\bottomrule
\end{tabular}
\end{table*}

\begin{table}[ht]
\small
\centering
\caption{Explicit Citation of the Provided Answer in the AE Condition.}
\label{tab:citation-rate}
\small
\setlength{\tabcolsep}{6pt}
\begin{tabular}{>{\raggedright\arraybackslash}p{5cm} >{\centering\arraybackslash}m{2cm}}
\toprule
\textbf{Models} & \textbf{Citation Rate} \\
\midrule
\makecell[l]{\deepseek} & 12.03\% \\
\makecell[l]{\oThree}  & 8.86\% \\
\makecell[l]{\oThreeMini} & 12.66\% \\
\makecell[l]{\oFourMini}  & 13.29\% \\
\makecell[l]{\qwq} & 6.96\% \\
\makecell[l]{\grokBeta}  & 38.61\% \\
\makecell[l]{\grokFast}  & 39.87\% \\
\makecell[l]{\geminiFlashnew}  & \makecell[c]{25.32\%} \\
\makecell[l]{\geminiPro}  & 19.62\% \\
\makecell[l]{\claude}  & 20.25\% \\
\midrule
\textbf{Avg. Rate} & \textbf{19.75\%} \\
\bottomrule
\end{tabular}
\end{table}

\section{Experiments}\label{sec:experiment}
This section presents our empirical investigation, structured around three interconnected inquiries designed to dissect the nature and strength of memory anchoring in LLMs:

\begin{itemize}
    \item Are LLMs primarily anchored to final answers or learned reasoning templates (Section~\ref{subsec:main_result}) ?

    \item If answers and reasoning templates conflict, which do LLMs prioritize? (Section~\ref{subsec:memory_conflict}) ?

    \item What is the depth of LLMs' memory anchoring to the dominant cue?  (Section~\ref{subsec:prompt_helps}) ?
\end{itemize}

\subsection{Where Is Memory Bound: Answer or Reasoning?}\label{subsec:main_result}

We evaluate model performance across five progressively constrained conditions that systematically reduce both explicit and implicit answer cues from the prompt. This design directly assesses the extent to which final answers, relative to reasoning templates, guide model behavior.

As shown in Table~\ref{tab:performance_comparison}, overall accuracy declines monotonically as answer visibility decreases. Under the AE condition, models achieve consistently high performance, with an average accuracy of 84.75\%. Top-tier models such as DeepSeek-R1 (93.04\%) and Claude 3.7 Sonnet (93.67\%) approach near-perfect scores. However, when the answer is only embedded in the reasoning chain (AER), accuracy drops to 76.96\%. This suggests that implicit cues within reasoning chains, while helpful, are still markedly weaker than explicitly provided answers. The decline becomes even more substantial in the AMR setting, where the answer token is masked. Here, the average accuracy drops sharply to just 57.85\%, indicating a strong reliance on the visible answer tokens. The most pronounced decline is observed in DeepSeek-R1, which drops from 84.18\% to 55.06\%, a nearly 30-point decrease, underscoring a heavy dependence on token-level answer visibility.

This vulnerability extends beyond explicit answer tokens, as reasoning chains often include paraphrased or derived sentences that implicitly imply the answer. Consequently, in ARR, excising such answer-related sentences further reduces accuracy to 46.46\%.  Finally, in the AF condition, where no answer or reasoning cues are provided, performance collapses to a 28.17\% baseline.

This trajectory indicates that LLM memory is predominantly anchored to final answers, rather than to the textual patterns of reasoning templates. Providing an explicit answer (AE) yields a substantial 56.58\% improvement over the AF baseline, far surpassing the 29.68\% gain observed when only a masked answer is embedded within a reasoning chain(AMR). The 26.90\% margin between AE and AMR highlights the disproportionate influence of explicit presented answers, confirming their dominant role in guiding model responses.

Beyond this primary finding, further analysis reveal additional aspects of LLM behavior. One striking observation is the model’s tendency to perform post-hoc rationalization—generating plausible reasoning chains to support an already-known answer. For instance, with the generated reasoning chains, models achieve 76.96\% accuracy. Yet, when the final answer token is masked (AMR), performance drops by 19.11\%, despite the reasoning chain being intact. This suggests that reasoning steps alone are insufficient for robust inference, and that much of the observed “reasoning” may reflect retrospective alignment rather than forward derivation. Such behavior may lead to an overestimation of LLMs' independent inferential capabilities.

\begin{table*}[ht]
\small
\centering
\caption{Analysis of Incorrect Answers with Correct Reasoning and Vice Versa.}
\label{tab:conflicting-cues}
\small
\setlength{\tabcolsep}{6pt}
\begin{tabular}{
  >{\raggedright\arraybackslash}p{5cm}
  >{\centering\arraybackslash}m{3cm}
  >{\centering\arraybackslash}m{3cm}
  >{\centering\arraybackslash}m{3cm}
}
\toprule
\textbf{Models} &
\makecell{\textbf{Wrong Answer} \\ \textbf{+ Wrong Reasoning}} &
\makecell{\textbf{Wrong Answer} \\ \textbf{+ Right Reasoning}} &
\makecell{\textbf{Right Answer} \\ \textbf{+ Wrong Reasoning}} \\
\midrule
\makecell[l]{\deepseek} & 6.96\% & 39.24\% & 52.53\% \\
\makecell[l]{\qwq} & 10.13\% & 40.51\% & 53.16\% \\
\makecell[l]{\grokBeta} & 7.59\% & 50.00\% & 57.59\% \\
\makecell[l]{\grokFast} & 6.33\% & 51.90\% & 55.06\% \\
\makecell[l]{\claude} & 9.49\% & 25.32\% & 64.56\% \\
\midrule
\textbf{Avg. Performance} & \textbf{8.10\% ($\pm 1.64\%$)} & \textbf{41.39\% ($\pm 10.59\%$)} & \textbf{56.58\% ($\pm 4.88\%$)} \\
\bottomrule
\end{tabular}
\end{table*}

Interestingly, Table~\ref{tab:citation-rate} details another notable behavior observed under the AE condition: the rates at which models explicitly cite the provided answer. Citation judgments are  conducted automatically using GPT-4o-1120 (see Appendix~\ref{subsec:citation_judge} for prompt details).  These citation rates are generally modest and exhibit considerable variation across different models, ranging, for instance, from 6.96\% (QwQ-32B) to 39.87\% (Grok-3 mini fast Beta-high). This finding suggests that even when a model clearly benefits from the inclusion of an explicit answer, it does not consistently acknowledge or integrate that answer into its surface-level output. Such behavior highlights a disconnect between what the model uses and what the model says it uses, raising questions about transparency and attribution in LLM-generated reasoning.

\subsection{Memory Preference Under Conflicts}\label{subsec:memory_conflict}

Building on the findings of LLM answer-anchoring (Section~\ref{subsec:main_result}), we further examine memory preference in scenarios where the final answers and reasoning chains conflict. Such conflicts leverage RoR‑Bench’s design, in which each problem is paired with its unmodified source question. The RoR problem comes with a newly defined correct answer and corresponding reasoning chain.  In contrast, the answer and reasoning chain from its corresponding original version serve as incorrect yet superficially similar cues for the modified task. This setup enables controlled semantic conflicts between answer-level and reasoning-level cues. We evaluate model behavior under three such configurations:

\begin{itemize}
\item Right Ans / Wrong Reasoning (RA/WR): the RoR problem's correct answer paired with the original problem's reasoning chain. 

\item  Wrong Ans / Right Reasoning (WA/RR): the original problem's answer paired with the RoR problem's own correct reasoning chain. 

\item Wrong Ans / Wrong Reasoning (WA/WR): a baseline using both the answer and reasoning derived from the original problem's solution.

\end{itemize}

As shown in Table~\ref{tab:conflicting-cues}, models perform poorly when both the answer and reasoning are incorrect (the WA/WR baseline), averaging only 8.10\% accuracy. In contrast, when cues conflict, models clearly prioritize the answer. On average, accuracy reaches 56.58\% when the answer is correct but the reasoning is flawed (RA/WR), substantially exceeding the 41.39\% observed when the reasoning is correct but the answer is deliberately misleading (WA/RR).  Claude 3.7 Sonnet further exemplifies this trend, with its accuracy reaching 64.56\% in the RA/WR setting, far exceeding its 25.32\% in WA/RR. These results highlight the dominant influence of answers, confirming that models perform better when the provided answer is correct even if the supporting reasoning is not.

These findings collectively confirm that LLMs tend to prioritize explicit answers over reasoning when faced with conflicting signals, a conclusion that strongly corroborates the answer-anchoring observations. However, the fact that RA/WR average performance (56.58\%) remains well below that of the AE condition (84.75\%) indicates that  flawed reasoning still markedly impairs performance, underscoring reasoning's essential—though secondary—role.

\begin{figure*}[ht]
\begin{center}
\includegraphics[width=1.0\textwidth, trim=0cm 0cm 0cm 0cm, clip]{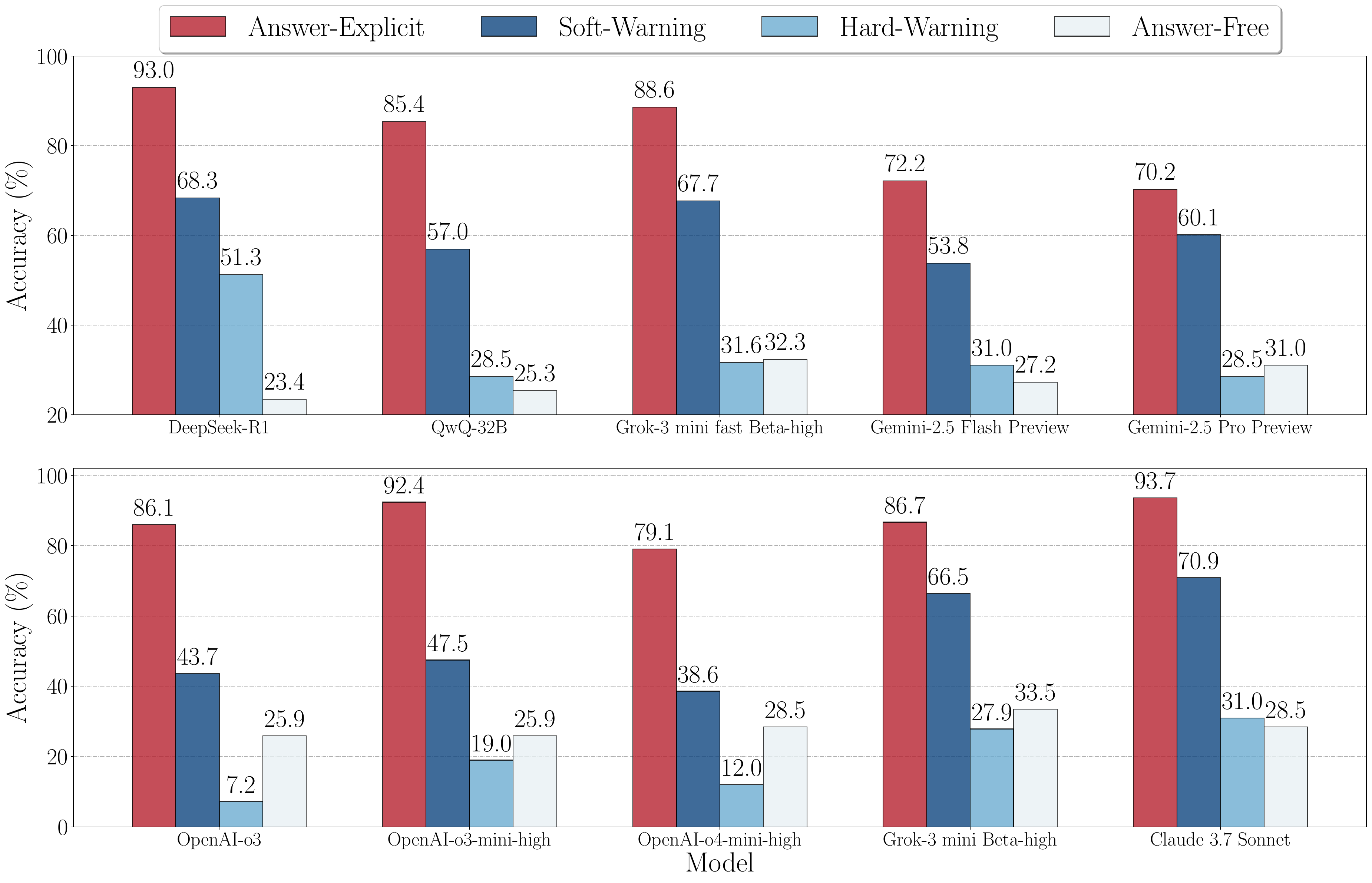}
\end{center}

\caption{Impact of Soft and Hard Warning prompts on LLM accuracy in an Answer-Explicit setting.}
\label{fig:prompt}
\end{figure*}

\subsection{Probing the Tenacity of Answer Anchoring }\label{subsec:prompt_helps}

To further probe the tenacity of LLMs' dependency on explicit answers, we investigate how 'warning prompts' affect reliance on such cues. Specifically, we implement two variants: a soft warning and a hard warning. The soft one states: "Please answer the following question carefully. Note: The reference answers may be incorrect and are for reference only. Please rely on your own independent reasoning to provide the answer that best fits the question." In contrast, the hard warning removes ambiguity by asserting that "The reference answers are incorrect." This setup allows us to assess whether LLMs can override known but misleading answer cues when prompted to distrust them.

Figure~\ref{fig:prompt} shows warning prompts consistently reduce accuracy relative to the AE baseline. Hard warnings are typically more impactful than soft ones, indicating that models are sensitive to warning intensity. However, responses vary considerably across models. For instance, OpenAI models exhibit steep declines, while Gemini models show greater resilience, especially under soft warnings. Notably, DeepSeek-R1 uniquely maintains 51.3\% accuracy even with a hard warning. 

This accuracy reduction, when models are warned against a correct answer, suggests a degree of instructability, as models attempt to heed warnings. Yet, the answer's influence remains tenacious, with models like DeepSeek-R1 and Claude 3.7 Sonnet under hard warning still outperform their AF scores. In contrast, others suffer sharper declines. For some, hard warnings push accuracy near or below AF levels, suggesting that conflicting cues can severely disrupt processing. These varied responses reflect how differently models anchor to answers when challenged.

Ultimately, these experiments with skepticism-inducing prompts underscore the remarkable tenacity of answer anchoring in LLMs. While models show some responsiveness to instructions intended to weaken this reliance, the influence of the provided answer remains substantial—even when explicitly discredited. That LLMs struggle to fully and consistently disengage from salient answer cues, including correct ones they are warned to distrust, highlights their fundamentally answer-centric behavior.  This reinforces the primary thesis that LLM memory and processing are predominantly bound to answers, revealing the significant extent of this dependency when directly challenged by such countervailing instructions.

\section{Conclusion}

In this work, we systematically investigate the nature of memory anchoring in LLMs when solving reasoning tasks. By manipulating the visibility of final answers within prompts, we uncover a profound and consistent pattern: LLM performance is predominantly anchored to the explicit presence of final answers rather than to the textual patterns of the reasoning steps themselves. 

Furthermore, we demonstrate that while LLMs can generate seemingly coherent reasoning when answers are provided, their ability to deduce correct answers solely from reasoning chains remains severely limited. These findings are reinforced by experiments showing LLMs' preference for explicit answers even when cues conflict, and by the tenacious nature of this answer dependency despite designed warnings to suppress it. These results suggest that the reasoning exhibited by LLMs may often be a form of post-hoc rationalization around a known or anticipated answer, rather than independent inference. This challenges common assumptions about LLM reasoning depth and underscores the urgent need to rethink how reasoning capabilities are evaluated.

\textbf{Limitations}
Our investigation primarily uses the text-only RoR-Bench dataset~\citep{yan2025recitation}, focusing on specific reasoning types. The generalizability of our findings to other domains, languages, or modalities (such as visual reasoning) warrants further exploration. Moreover, our core experimental manipulation—providing the answer with the input prompt—is an artificial setup. While diagnostically powerful for isolating variables, its divergence from typical real-world LLM interactions suggests that other probing techniques might reveal additional facets of model reasoning.

\newpage


\newpage
~
\newpage
\appendix
\newpage

\section{Appendix}
\label{sec:appendix}
\subsection{Prompt for the Judge}\label{subsec:judge}

\begin{center}
\includegraphics[width=1.0\textwidth, trim=0cm 0cm 0cm 0cm, clip]{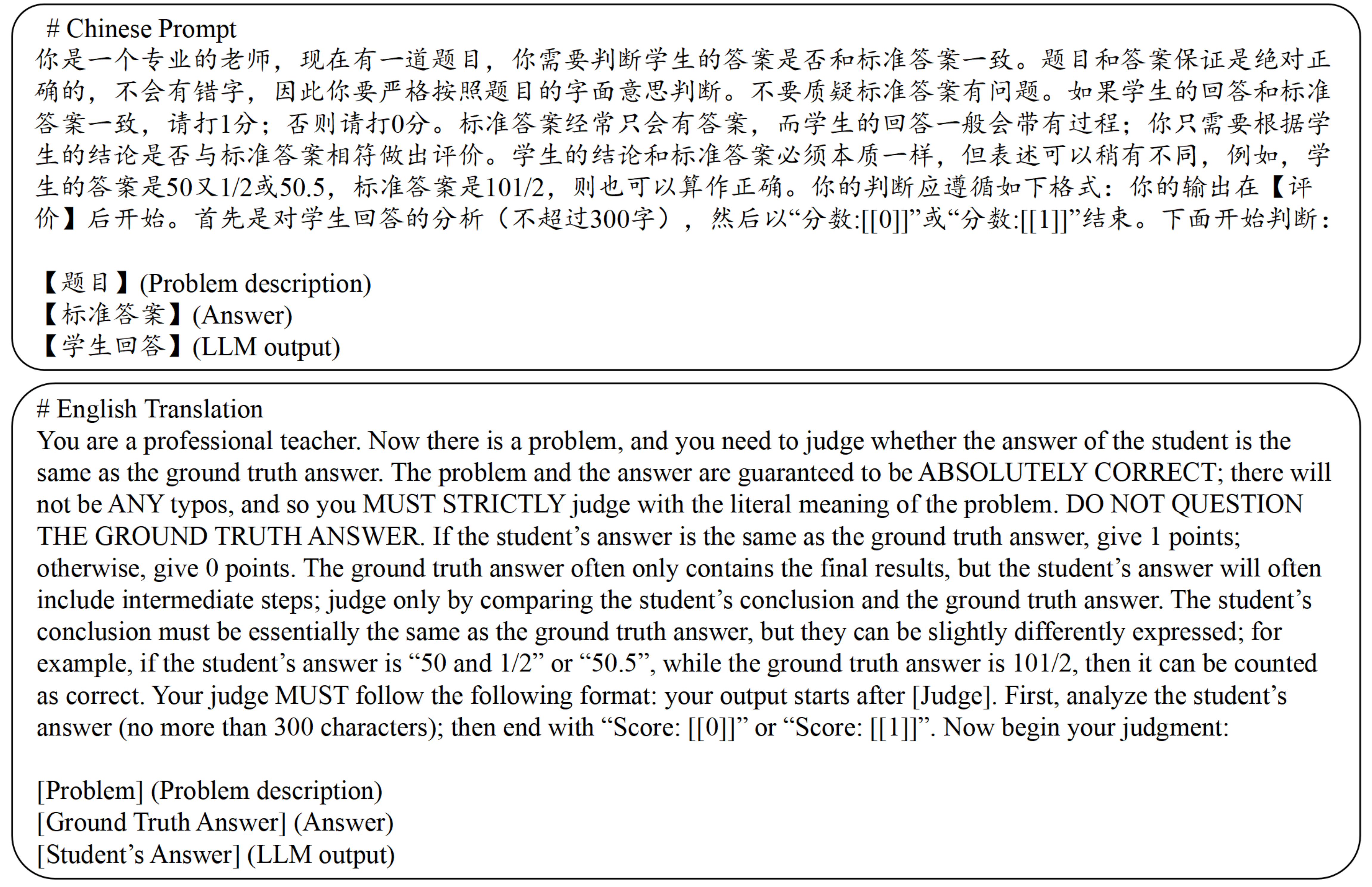}
\label{fig:prompt_judge}
\end{center}

For all experiments, we consistently used the same Chinese evaluation prompt for GPT-4o-1120 as the judge. For clarity, English translations accompany each presented case. Crucially, we adopt the prompt design from~\citep{yan2025recitation}. Please refer to Figure~\ref{fig:prompt_judge} for the specific prompt used. This design emphasizes the directive "do not question the ground truth answer." The inclusion of this specific instruction is vital, as its absence leads the LLMs to critique the problem formulation itself, rather than evaluate the student’s response, even when the ground truth is clearly provided.

\newpage
~
\newpage
\subsection{Judge Prompt for Answer Citation}\label{subsec:citation_judge}
\begin{center}
\includegraphics[width=0.9\textwidth]{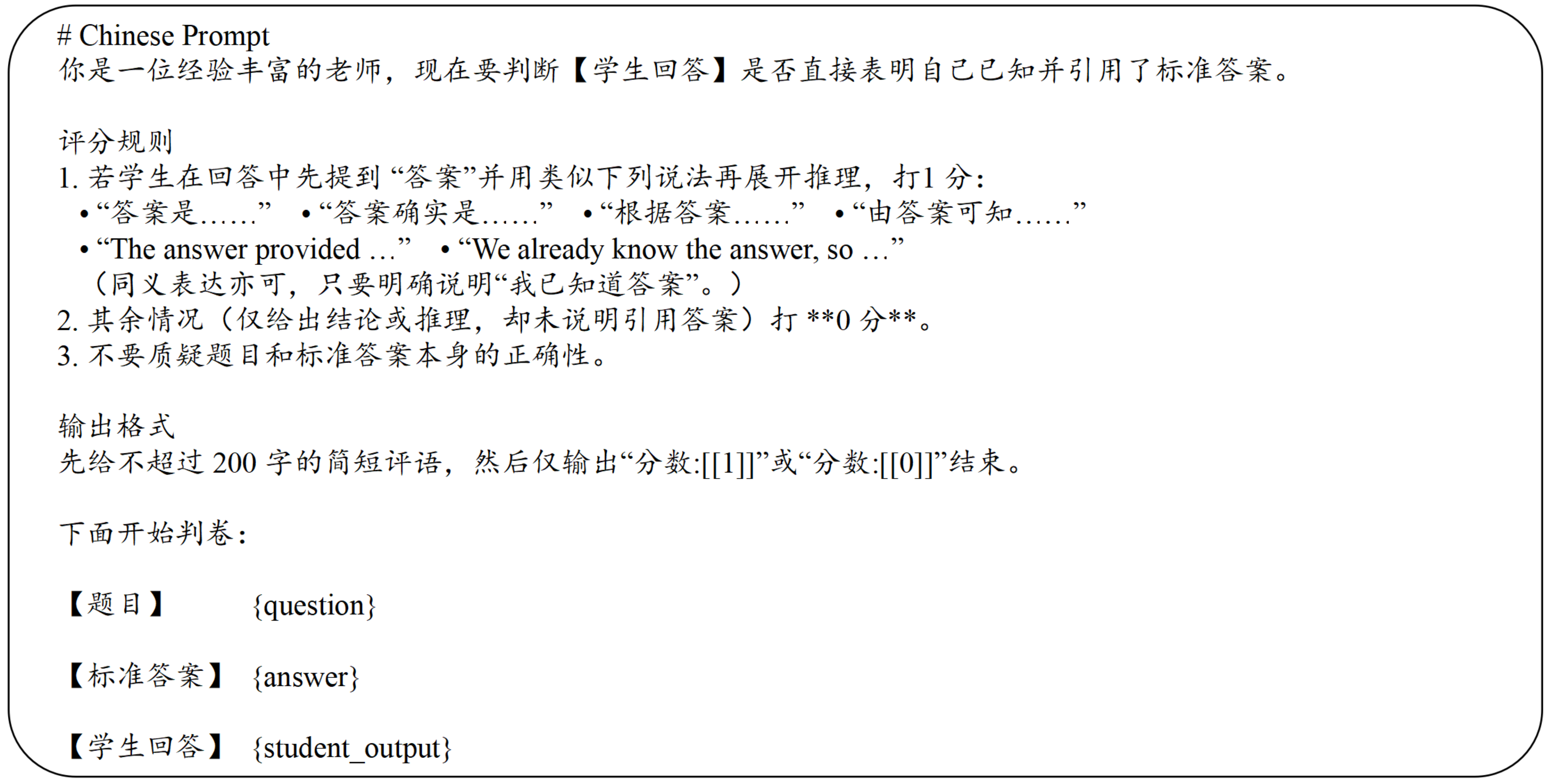}
\label{fig:prompt_chinese}
\end{center}

\vspace{-3em}  

\begin{center}
\includegraphics[width=0.9\textwidth]{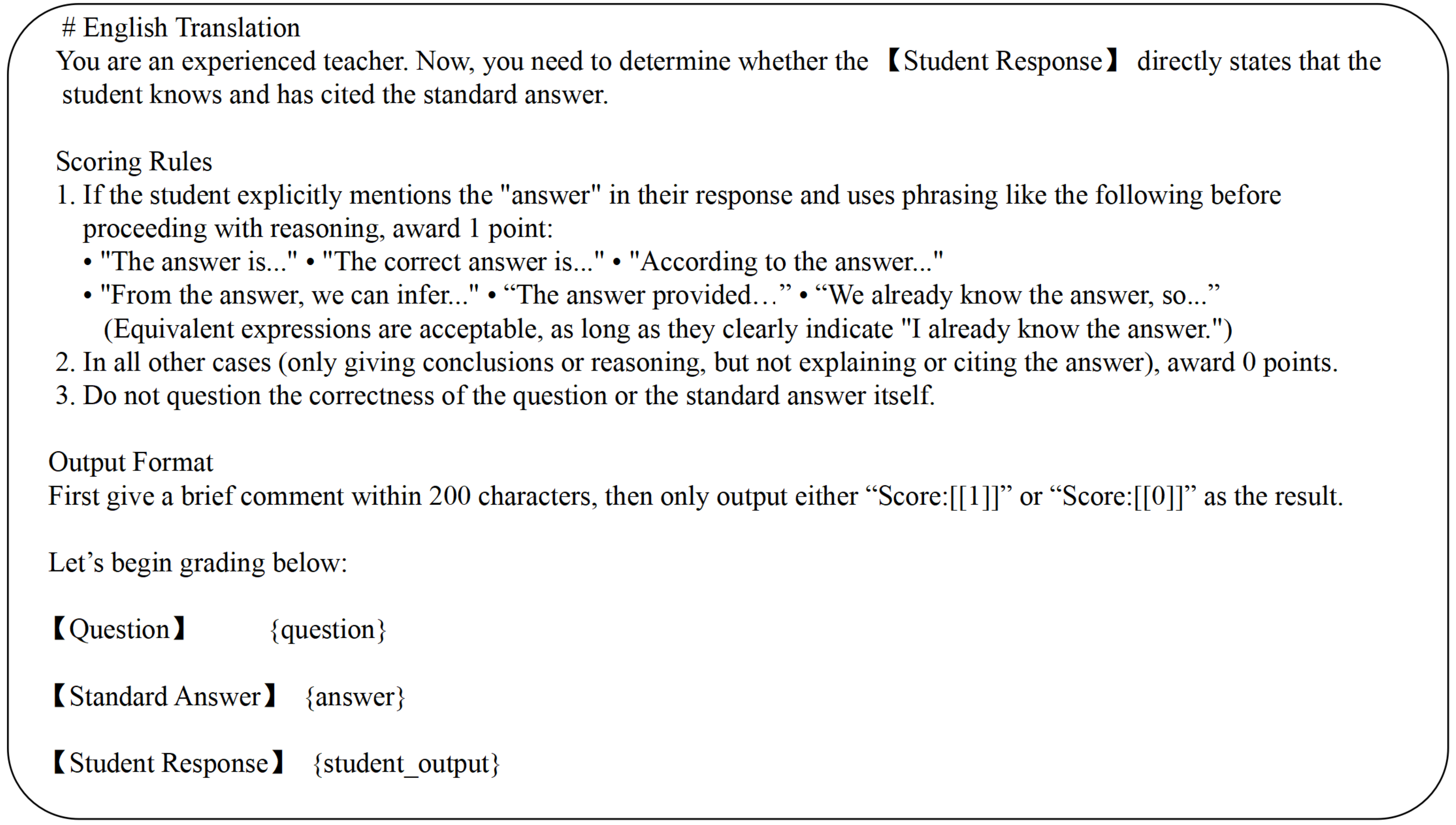}
\label{fig:prompt_english}
\end{center}

We use a dedicated evaluation prompt for GPT-4o-1120 to judge whether a model's response explicitly cites the provided standard answer.  As the original prompts and answers are in Chinese, we use a Chinese prompt for judging and provide the corresponding English translation alongside. 

This prompt plays a crucial role in identifying whether the model is merely solving the problem or actively acknowledging the given answer. It emphasizes the detection of phrases such as “the answer is...” or “according to the answer...”, and only assigns credit when citation is made explicit. Without this explicit checking prompt, LLMs often produce valid responses without ever referencing the known answer, making it difficult to distinguish true citation behavior from general correctness.

\newpage
~
\newpage

\subsection{Prompt for the Answer Masking}\label{subsec:mask_prompt}
\begin{center}
\includegraphics[width=0.9\textwidth]{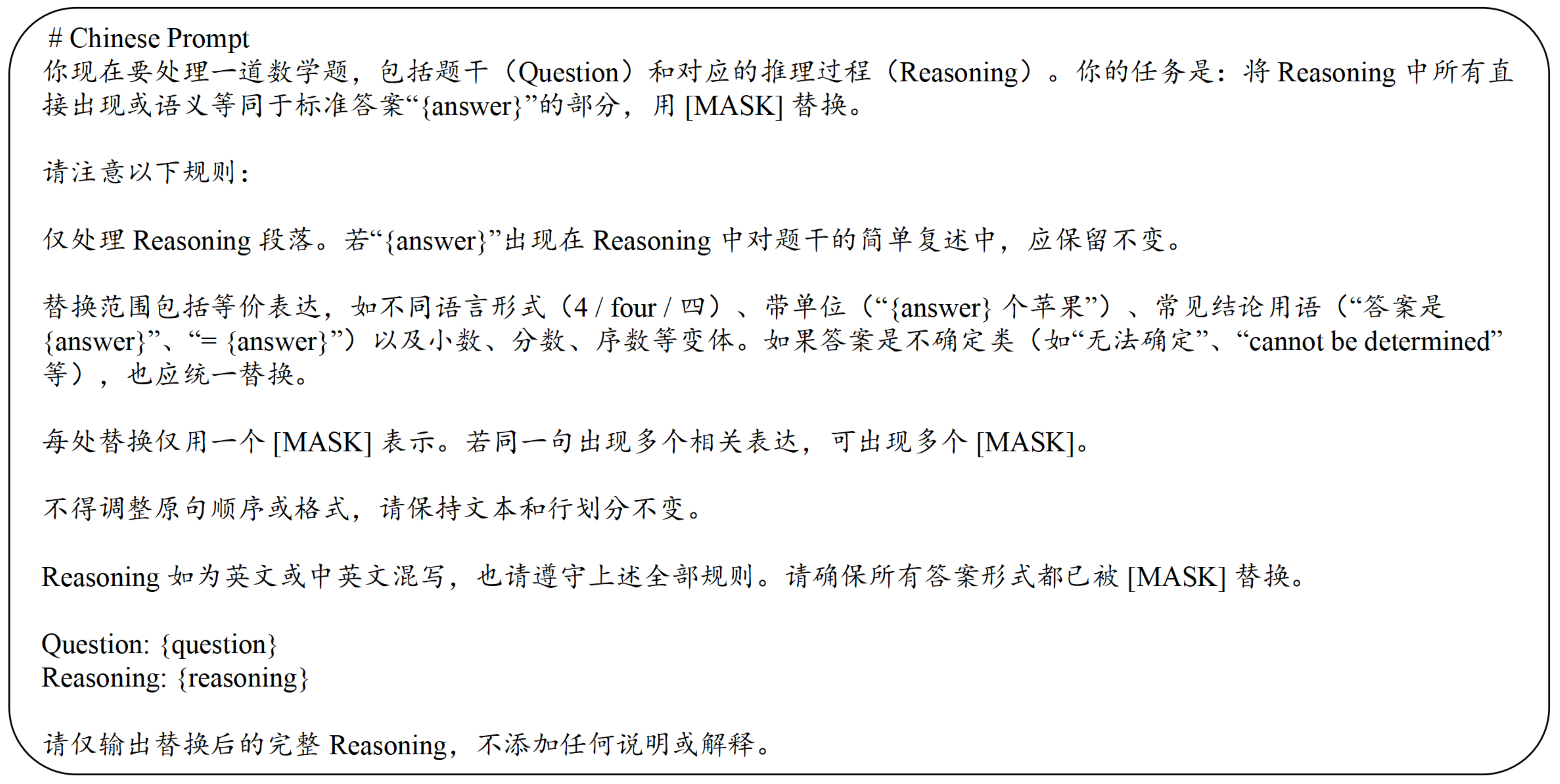}
\label{fig:prompt_chinese}
\end{center}

\vspace{-3.5em}  

\begin{center}
\includegraphics[width=0.9\textwidth]{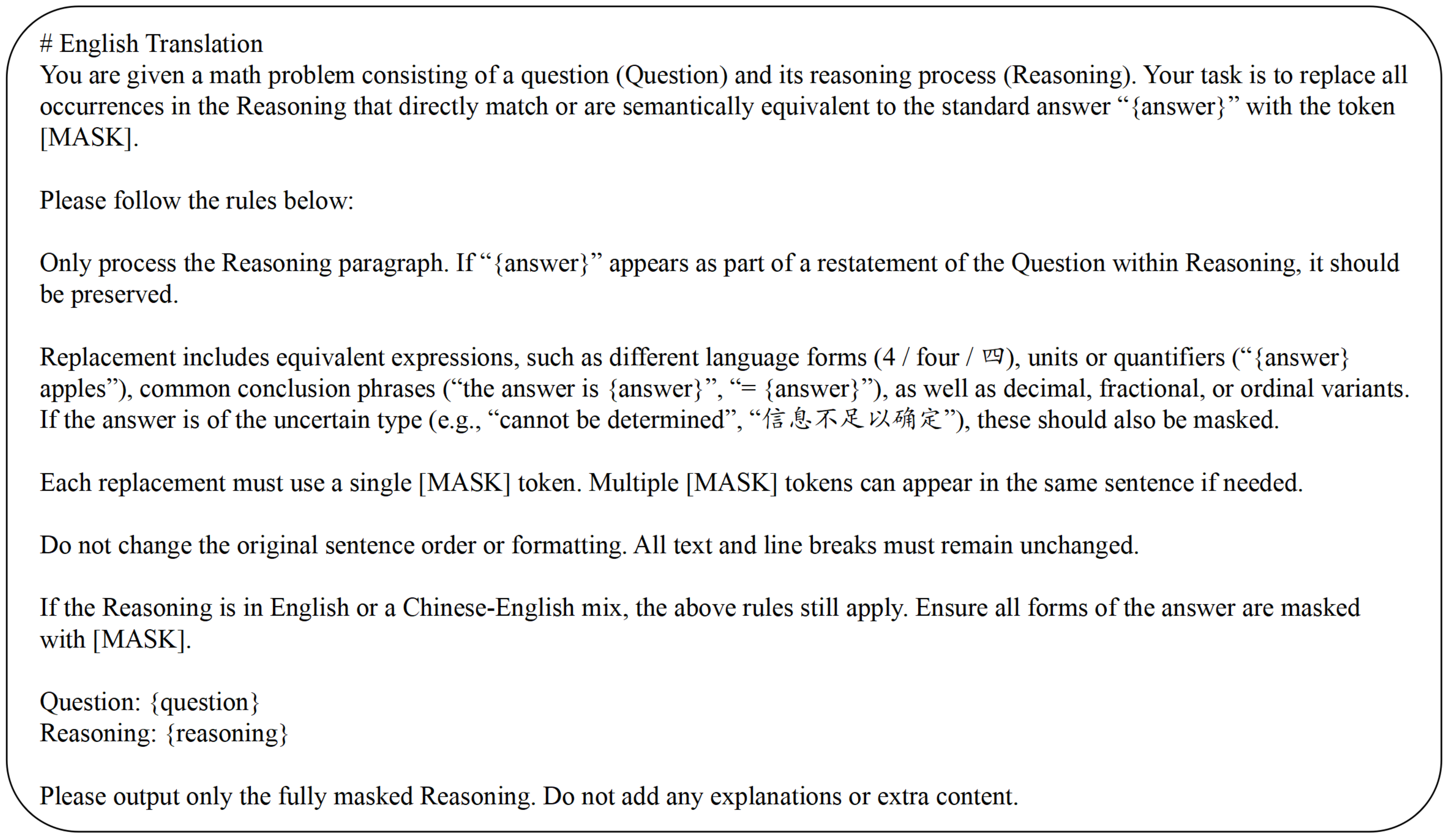}
\label{fig:prompt_english}
\end{center}

To support our masking experiments, a dedicated prompt using  GPT-4o-1120 systematically eliminates all explicit and semantically equivalent references to the answer from reasoning chains. This process targets a comprehensive range of answer expressions—covering diverse linguistic, numerical, unit-quantified, ordinal, conclusive, and uncertainty forms—replacing each with \texttt{[MASK]}, potentially multiple times per sentence if warranted. This automated masking preserves the structure of the original reasoning while fully suppressing answer-related content. Such precise control is essential for preventing subtle answer leakage and ensuring the integrity of answer-agnostic evaluations.



\end{document}